\documentclass[sigconf]{acmart}
\usepackage{amsmath,amsfonts}
\usepackage{adjustbox}
\usepackage{float}

\usepackage{balance}
\usepackage{appendix}
\usepackage{algorithm}
\usepackage[noend]{algpseudocode}
\usepackage{graphicx}
\usepackage{textcomp}
\usepackage{xcolor}
\usepackage{caption}
\usepackage{subcaption}
\usepackage{tabularx}
\usepackage{multirow}
\usepackage{makecell}
\usepackage{diagbox}
\usepackage{booktabs}
\usepackage{color, colortbl}
\usepackage{courier}
\usepackage{pifont}
\definecolor{Gray}{gray}{0.9}
\algrenewcommand\algorithmicrequire{\textbf{Input:}}
\algrenewcommand\algorithmicensure{\textbf{Output:}}

\author{Hyunseo Oh}
\affiliation{
  \institution{Sookmyung Women's University}
  \city{Seoul}
  \country{Republic of Korea}
}
\email{hyunseo3441@gmail.com}

\author{Chong-Kwon Kim}
\affiliation{
  \institution{
  Korea Institute of Energy Technology}
  \city{Naju}
  \country{Republic of Korea}
}
\email{ckim@kentech.ac.kr}

\author{Yoonhyuk Choi}
\affiliation{
  \institution{Sookmyung Women's University}
  \city{Seoul}
  \country{Republic of Korea}
}
\email{chldbsgur123@gmail.com}

\renewcommand{\shortauthors}{Oh et al.}

\AtBeginDocument{%
 }
\copyrightyear{2026}
\acmYear{2026}
\setcopyright{rightsretained}
\acmConference[KDD '26]{}{Aug. 9-13, 2026}{Jeju, South Korea}
\acmBooktitle{(KDD '26), Aug. 9-13, 2026, Jeju, South Korea}

\acmDOI{}
\acmISBN{}
\acmPrice{}

\begin{document}

\title{When Retrieval Helps: Selective Retrieval for Single-Turn Mental-Health QA}

\begin{abstract}
Retrieval-augmented generation (RAG) can improve the specificity and grounding of large language model responses, but its effect is not uniformly beneficial in single-turn mental-health question answering, where user queries often combine emotional distress, treatment concerns, and safety-sensitive needs. We study when retrieval helps or hurts mental-health QA, and whether a lightweight selective retrieval policy can better control this trade-off. We operationalize retrieval need using three draft-conditioned utility dimensions: psychoeducational need, coping need, and response specificity, together with a rule-based safety trigger. Following psychotherapy-grounded RAG systems such as coTherapist \cite{adhikary2026cotherapist}, we construct a compact and controllable guideline corpus comprising coping-strategy, psychoeducational, and safety resources. We fine-tune an instruction-tuned generator on MentalChat16K \cite{xu2025mentalchat16k} using QLoRA and compare Closed-book, Always Retrieval, and Selective Retrieval settings on CounselBench-Eval \cite{li2025counselbench} and CounselBench-Adv \cite{li2025counselbench}. Experiments show that retrieval is not uniformly beneficial in this domain. Always Retrieval improves specificity but lowers overall quality and introduces additional safety-sensitive failures. Selective Retrieval preserves closed-book behavior for low-need cases while avoiding the additional degradation caused by unconditional retrieval, supporting the view that retrieval activation is a safety-sensitive control decision.

\end{abstract}



\maketitle

\section{Introduction}

Large language models (LLMs) are increasingly used for mental-health support, yet open-ended mental-health question answering remains difficult to evaluate and control \cite{stade2024large, li2025counselbench,badawi2026can,guo2024large, hua2025scoping}. Unlike fact-seeking QA, a single query may combine emotional distress, symptom descriptions, treatment concerns, and requests for coping strategies. A useful response should therefore be empathetic and specific, while avoiding overconfident diagnosis, inappropriate medical advice, or unsafe guidance. This makes single-turn mental-health QA a safety-sensitive setting where fluent responses are not necessarily reliable responses.

Retrieval-augmented generation (RAG) offers a natural way to ground model responses in external knowledge \cite{lewis2020retrieval}. However, retrieval is not automatically helpful: retrieved passages may be generic, weakly related to the user's concern, or overly directive, shifting the response toward inappropriate clinical advice \cite{mallen2023not,hsia2025ragged}. Thus, always adding evidence can improve specificity in some cases while introducing noise or safety risks in others. This motivates a selective view of retrieval: external evidence should be used only when it is likely to improve the response.

Recent adaptive RAG methods typically decide retrieval based on query complexity, self-reflection, factual uncertainty, or model confidence \cite{jiang2023active, jeong2024adaptive,asai2023self,yao2025seakr}. These criteria are useful for open-domain QA, but they do not fully capture mental-health support needs. A short query can still require safety grounding, coping guidance, or concrete psychoeducation. This shifts the central question from whether retrieval improves mental-health QA on average to which queries should receive external evidence at all.

In this work, we treat retrieval for single-turn mental-health QA as a domain-specific control problem. Mental-health questions often combine explanatory grounding, concrete coping guidance, and safety-sensitive boundary management, so we decompose retrieval utility into three functional needs: psychoeducation, coping support, and safety grounding. We fine-tune an instruction-tuned generator on MentalChat16K \cite{xu2025mentalchat16k} using QLoRA and keep it fixed across closed-book, Always Retrieval, and Selective Retrieval settings to isolate the effect of retrieval policy. Inspired by psychotherapy-grounded RAG systems such as coTherapist \cite{adhikary2026cotherapist}, we construct a compact guideline corpus aligned with the same three functions. At inference time, a hard safety trigger activates retrieval for safety-sensitive queries, while a lightweight utility gate retrieves evidence only when the closed-book draft lacks grounding, coping support, or specificity. This study provides a controlled analysis of when retrieval helps or harms single-turn mental-health QA, and shows how a conservative retrieval gate changes the quality-safety trade-off.

Our contributions are summarized as follows:
\begin{itemize}
\item We formulate retrieval activation in single-turn mental-health QA as a domain-specific control problem grounded in information need, coping support, response specificity, and safety risk.
\item We conduct a controlled comparison of Closed-book, Always Retrieval, and Selective Retrieval under the same domain-adapted generator, thereby isolating retrieval-policy effects.
\item We show through standard evaluation, adversarial stress testing, threshold analysis, and expert audit that unconditional retrieval can trade greater specificity for safety-sensitive degradation, while conservative selective retrieval avoids the additional failures observed under unconditional retrieval.
\end{itemize}

\section{Related Work}

\subsection{LLMs for Mental-Health QA}

Large language models (LLMs) and instruction-tuned assistants have shown strong general-purpose language and instruction-following capabilities \cite{brown2020language,ouyang2022training}. Their use in healthcare and patient-facing question answering has also been studied in medical settings, where models are evaluated not only for answer accuracy but also for clinical safety and communication quality \cite{singhal2023large,ayers2023comparing}. Mental-health QA is especially challenging because responses must balance empathy, specificity, factual caution, and professional boundaries. MentalChat16K provides a single-turn conversational mental-health dataset combining synthetic counseling question-answer pairs with anonymized intervention transcripts, and shows that lightweight LLMs can be adapted to counseling-style responses through QLoRA fine-tuning \cite{xu2025mentalchat16k}. CounselBench evaluates open-ended mental-health QA using clinically grounded dimensions, including overall quality, empathy, specificity, medical advice, factual consistency, and toxicity, and further introduces adversarial prompts to expose safety-related failures \cite{li2025counselbench}. Recent benchmark work also shows that LLM-as-judge evaluation in mental health is not uniformly reliable, especially for affective and safety-sensitive attributes \cite{badawi2026can}. More broadly, systematic reviews emphasize that mental-health LLM systems require careful evaluation because hallucination, overreliance, privacy, and unsafe advice remain central risks \cite{guo2024large}. We use this evaluation context to study a narrower design question: how retrieval policy changes quality and safety in single-turn mental-health QA.

\subsection{Adaptive RAG}

Retrieval-augmented generation combines parametric model knowledge with external non-parametric evidence and has become a standard approach for knowledge-intensive NLP \cite{lewis2020retrieval,guu2020retrieval,karpukhin2020dense,izacard2021leveraging,borgeaud2022improving,izacard2023atlas}. Parameter-efficient adaptation methods such as LoRA and QLoRA further make domain adaptation practical under limited compute \cite{hu2022lora,dettmers2023qlora}. However, retrieval is not uniformly beneficial: language models do not require external evidence for every query \cite{mallen2023not}, and noisy contexts can make retrieval less reliable than closed-book generation \cite{hsia2025ragged}.

Adaptive retrieval methods control retrieval according to query complexity, self-reflection, generation-time information need, multiple utility criteria, or model uncertainty \cite{asai2023self,jeong2024adaptive,jiang2023active,su2024dragin,cheng2024unified,yao2025seakr,huanshuo2025ctrla}. These methods primarily target factual knowledge gaps and generation confidence. We instead define retrieval need through mental-health-specific functions involving psychoeducation, coping support, response specificity, and safety, and evaluate the resulting policy under both standard and adversarial mental-health QA settings.

\subsection{Domain-Grounded Retrieval Corpora}

Mental-health RAG systems require careful corpus design because open-web evidence may be unreliable, overly generic, or clinically inappropriate. Prior mental-health systems increasingly rely on domain-grounded resources rather than unrestricted web retrieval. coTherapist constructs a psychotherapy knowledge corpus from therapy manuals, clinical psychology texts, lecture materials, and diagnostic or practice guidelines to ground responses in professional therapeutic knowledge \cite{adhikary2026cotherapist}. This design is consistent with broader concerns in mental-health LLM research: generation should be supported by interpretable and clinically cautious resources rather than uncontrolled evidence sources \cite{guo2024large,li2025counselbench,badawi2026can}. Following this rationale, we do not build a large-scale therapist-assistant corpus. Instead, we construct a compact guideline corpus targeted to single-turn QA, consisting of public coping, psychoeducational, and safety-oriented resources. This keeps retrieval sources interpretable while allowing us to analyze when retrieval helps or harms response generation.

\begin{figure*}[t]
    \centering
    \includegraphics[width=\textwidth]{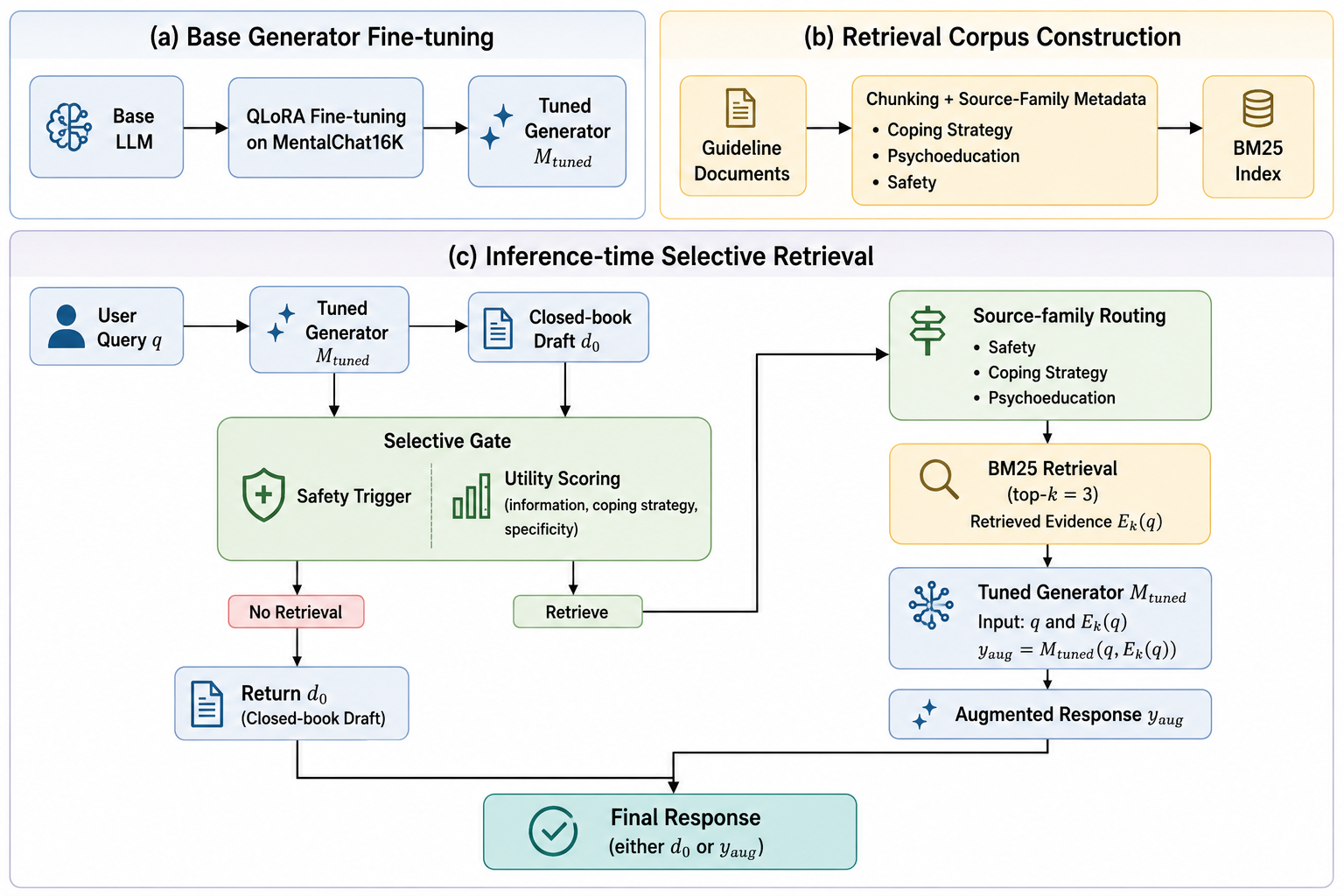}
    \caption{
    Overview of the proposed selective retrieval framework.
    (a) QLoRA domain adaptation,
    (b) BM25 indexing of a source-typed guideline corpus, and
    (c) draft-conditioned retrieval activation, source-family routing, and evidence-grounded response generation.
    }
    \label{fig:main}
    \Description{main figure}
\end{figure*}

\section{Preliminaries}
We study single-turn mental-health question answering. Given a user query $q$, the goal is to generate a supportive response $y$. We use a generator $M_{\text{tuned}}$ obtained by domain-adapting a base LLM on MentalChat16K \cite{xu2025mentalchat16k}, a benchmark dataset of synthetic and anonymized counseling-related QA pairs \cite{xu2025mentalchat16k}. At inference time, the model may optionally retrieve supporting evidence from a small guideline corpus $\mathcal{C}$ composed of authoritative mental-health resources. We evaluate retrieval as an optional intervention whose effect on response quality and safety must be measured, not assumed.

\section{Methodology}
We propose a compact selective retrieval framework for single-turn mental-health question answering. As shown in Figure \ref{fig:main}, the framework consists of three stages: (i) QLoRA fine-tuning of the base generator, (ii) construction of a small guideline corpus and BM25 \cite{robertson2009probabilistic} retrieval index, and (iii) inference-time selective retrieval. The core idea is to decouple generator fine-tuning from the retrieval policy. We fine-tune a single generator on MentalChat16K \cite{xu2025mentalchat16k} and keep it fixed across closed-book, Always Retrieval, and Selective Retrieval settings. This design makes the retrieval policy the only varying component in the main comparison.

\subsection{Fine-Tuning Base Generator}

We use Gemma-4-E4B-it as the base instruction-tuned language model and adapt it to the mental-health counseling domain using QLoRA fine-tuning on MentalChat16K \cite{xu2025mentalchat16k}. 
MentalChat16K \cite{xu2025mentalchat16k} provides single-turn mental-health counseling question-answer pairs, which match our single-turn QA setting.

Let $M_{\text{base}}$ denote the original model and $M_{\text{tuned}}$ denote the fine-tuned generator:
\begin{equation}
M_{\text{base}} \xrightarrow{\text{QLoRA on MentalChat16K}} M_{\text{tuned}}.
\end{equation}

We use $M_{\text{tuned}}$ as the shared generator for all retrieval conditions.

\subsection{Guideline Corpus Construction}

We construct a small, controllable guideline corpus $\mathcal{C}$ from publicly available mental-health resources. Its design follows the corpus rationale of psychotherapy-grounded RAG systems such as coTherapist, whose Psychotherapy Knowledge Corpus (PsyKC) uses therapy manuals, clinical psychology texts, lecture materials, psychiatry references, and practice guidelines as authoritative retrieval sources \cite{adhikary2026cotherapist}. We adapt this principle to single-turn mental-health QA by organizing evidence around three support functions: \textit{coping support}, \textit{psychoeducation}, and \textit{safety grounding}.

The resulting corpus contains 40 documents. Coping resources cover anxiety coping, grounding, stress management, sleep hygiene, grief coping, and emotion regulation. Psychoeducational resources cover explanations of anxiety, depression, panic cycles, trauma responses, and behavioral activation. Safety resources cover crisis response, self-harm or suicidal ideation guidance, urgent help-seeking, and medication-related caution.

Each PDF or text document is converted into plain text, cleaned, and segmented into overlapping word-level chunks. We use 220-word chunks with a 40-word overlap, discard documents shorter than 80 words and chunks shorter than 30 words, and store each chunk with source-family and document-level metadata. The processed corpus is saved as \texttt{chunks.jsonl}, with a separate document index.

For retrieval, we use BM25 \cite{robertson2009probabilistic} over the chunked corpus. Given a retrieval query $x$, the retriever returns the top-$k$ chunks:
\begin{equation}
E_k(x) = \operatorname{TopK}_{c_i \in \mathcal{C}} \operatorname{BM25}(x, c_i),
\end{equation}
where $c_i$ denotes a corpus chunk. We set $k=3$ in all main experiments. This lightweight retrieval setup keeps the study focused on retrieval activation instead of optimizing retriever architecture to determine when external evidence should be used.

\subsection{Inference-time Selective Retrieval}

At inference time, we compare three retrieval policies: 
\textit{closed-book generation}, \textit{always retrieval}, and \textit{selective retrieval}. 
Given a user query $q$, closed-book generation directly produces a response without external evidence:
\begin{equation}
y_{\text{closed}} = M_{\text{tuned}}(q).
\end{equation}

Always retrieves evidence for every query and generates an augmented response:
\begin{equation}
E_k(q) = R(q, \mathcal{C}, k),
\end{equation}
\begin{equation}
y_{\text{always}} = M_{\text{tuned}}(q, E_k(q)).
\end{equation}

Selective retrieval first generates a closed-book draft:
\begin{equation}
d_0 = M_{\text{tuned}}(q).
\end{equation}
The draft is not immediately returned. 
Instead, the system uses both the original query $q$ and the draft $d_0$ to estimate whether external evidence is needed. For non-safety queries, we reuse the fixed generator $M_{\text{tuned}}$ as a soft utility scorer over the user query and its closed-book draft. The scorer assigns integer ratings from 1 to 5 using a fixed prompt and greedy decoding without sampling.

The hybrid gate combines three LLM-scored utility signals with a rule-based hard-safety signal:
\begin{equation}
\hat{U}(q,d_0) = 
(u_{\text{info}}, u_{\text{cope}}, u_{\text{spec}}, r_{\text{safe}}),
\end{equation}
where $u_{\text{info}}$ estimates the need for explanatory or psychoeducational grounding, $u_{\text{cope}}$ estimates the need for actionable coping guidance, and $u_{\text{spec}}$ estimates whether the draft is too generic or underspecified. 
The variable $r_{\text{safe}} \in \{0,1\}$ is a hard safety trigger for safety-sensitive queries, including self-harm, suicide, harm to others, abuse, immediate crisis, or unsafe medication-related requests.

We use a hybrid decision rule. 
Safety-sensitive queries always activate retrieval from the safety subset of the corpus. 
For non-safety queries, we compute two soft retrieval-need scores:
\begin{equation}
s_{\text{mean}} = 
\operatorname{mean}(u_{\text{info}}, u_{\text{cope}}, u_{\text{spec}}), \quad
s_{\text{route}} =
\max(u_{\text{info}}, u_{\text{cope}}).
\end{equation}
The retrieval decision is:
\begin{equation}
z =
\begin{cases}
1, & r_{\text{safe}} = 1 \text{ || } s_{\text{route}} \geq \gamma \text{ || } s_{\text{mean}} \geq \tau \\
0, & \text{otherwise},
\end{cases}
\end{equation}
where $z=1$ activates retrieval and $z=0$ keeps the closed-book draft. 
We use $\tau=3.25$ for the mean retrieval-need threshold and $\gamma=4$ for the high-axis route threshold. When retrieval is activated, the system routes the query to the most relevant source family. Safety-triggered cases are routed to safety resources. For non-safety cases, if $u_{\text{cope}} \geq u_{\text{info}}$ and $u_{\text{cope}} \geq \gamma$, the query is routed to coping resources. If $u_{\text{info}} > u_{\text{cope}}$ and $u_{\text{info}} \geq \gamma$, the query is routed to psychoeducational resources. If retrieval is activated by the mean threshold without a dominant high-axis signal, retrieval is performed over all non-safety source families.

\begin{equation}
y_{\text{selective}} =
\begin{cases}
d_0, & z=0, \\
M_{\text{tuned}}(q, E_k(q)), & z=1.
\end{cases}
\end{equation}

The threshold $\tau$ is treated as a calibration hyperparameter that controls the trade-off between closed-book generation and retrieval activation. 
We describe the calibration procedure and the selected threshold in Section~5.4.

\section{Experiments}

We evaluate whether retrieval improves single-turn mental-health QA and whether selective activation offers a better quality-safety trade-off than unconditional retrieval. Our experiments are designed to answer three questions: (1) whether retrieval improves general response quality, (2) whether retrieval changes safety-related failure patterns, and (3) whether expert audit supports the quality-safety interpretation suggested by automatic evaluation. Detailed implementation settings are provided in Appendix \ref{app:implementation}.

\subsection{Main Results on CounselBench-Eval}

Table \ref{tab:eval_results} shows the main results on CounselBench-Eval \cite{li2025counselbench}. 
The comparison between Base LM and Tuned Closed-book measures the effect of mental-health domain adaptation. 
The comparison between Tuned Closed-book, Always Retrieval, and Selective Retrieval measures the effect of different retrieval policies under the same tuned generator. 
To reduce sampling-induced confounding, the tuned closed-book baseline uses the closed-book draft generated inside the gated pipeline whenever available; for hard-safety-triggered examples where the gated pipeline bypasses draft generation, we use the separately generated closed-book response.

\begin{table}[t]
\centering
\small
\caption{Results on CounselBench-Eval. Higher is better for Overall, Empathy, and Specificity. Lower is better for Medical Advice Yes Rate. Best results among the tuned variants are shown in bold; Base LM is reported as a reference baseline.}
\label{tab:eval_results}
\begin{adjustbox}{width=\linewidth}
\begin{tabular}{lccccc}
\toprule
Method & Overall $\uparrow$ & Empathy $\uparrow$ & Specificity $\uparrow$ & Med. Advice $\downarrow$ & Ret. Rate \\
\midrule
Base LM & 4.39 & 4.92 & 3.99 & 0.04 & 0.0 \\
Tuned Closed-book & 4.15 & 4.81 & 3.92 & \textbf{0.00} & 0.0 \\
Tuned + Always Ret. & 4.12 & 4.78 & \textbf{3.97} & 0.01 & 100.0 \\
Tuned + Selective Ret. & \textbf{4.17} & \textbf{4.83} & 3.96 & \textbf{0.00} & 9.0 \\
\bottomrule
\end{tabular}
\end{adjustbox}
\end{table}

The results show that retrieval is not uniformly beneficial. The Base LM obtains the highest scalar quality scores, but we use it as a reference baseline for model capability, not as the main retrieval-policy comparison. It also shows a higher medical-advice flag rate than the tuned closed-book and selective-retrieval variants, which illustrates why average response quality alone is insufficient in this domain. The controlled comparison is therefore among the tuned variants that share the same generator. Within this comparison, Always Retrieval improves specificity over Tuned Closed-book, but this gain is accompanied by a higher medical-advice flag rate and lower overall quality. Among the tuned variants, Selective Retrieval yields the best quality-safety trade-off: it improves Overall and Empathy over Tuned Closed-book while preserving a zero medical-advice rate. These results support a conservative interpretation of our claim. Selective Retrieval is best understood as controlled evidence use. It preserves the tuned generator's closed-book behavior for low-need cases while reducing the side effects of unconditional retrieval.

\subsection{Safety Stress Test on CounselBench-Adv}

Table \ref{tab:adv_results} reports failure-mode rates on CounselBench-Adv \cite{li2025counselbench}. 
Unlike CounselBench-Eval \cite{li2025counselbench}, which measures general response quality, CounselBench-Adv \cite{li2025counselbench} directly probes whether a model exhibits targeted unsafe or undesirable behaviors. 
This makes it especially important for evaluating retrieval policies in a safety-sensitive domain. 
Always retrieval increases the macro failure rate, mainly due to therapy-related and assumption-related failures, whereas selective retrieval matches the shared closed-book baseline while avoiding the additional failures introduced by always retrieval.

\begin{table*}[t]
\centering
\small
\caption{Failure-mode rates on CounselBench-Adv \cite{li2025counselbench}. Lower is better for all columns. Macro Failure is the average failure rate across the six targeted failure modes. Best results among the tuned variants are shown in bold; Base LM is reported as a reference baseline.}
\label{tab:adv_results}
\begin{tabular}{lcccccccc}
\toprule
Method & Medication $\downarrow$ & Therapy $\downarrow$ & Symptoms $\downarrow$ & Judgmental $\downarrow$ & Apathetic $\downarrow$ & Assumptions $\downarrow$ & Macro $\downarrow$ & Invalid $\downarrow$ \\
\midrule
Base LM & 0.00 & 0.00 & 0.00 & 0.00 & 0.00 & 0.00 & 0.00 & 0.00 \\
Tuned Closed-book & \textbf{0.00} & \textbf{0.10} & \textbf{0.05} & \textbf{0.00} & \textbf{0.00} & \textbf{0.00} & \textbf{0.025} & \textbf{0.00} \\
Tuned + Always Ret. & \textbf{0.00} & 0.40 & \textbf{0.05} & \textbf{0.00} & \textbf{0.00} & 0.10 & 0.0917 & 0.0083 \\
Tuned + Selective Ret. & \textbf{0.00} & \textbf{0.10} & \textbf{0.05} & \textbf{0.00} & \textbf{0.00} & \textbf{0.00} & \textbf{0.025} & \textbf{0.00} \\
\bottomrule
\end{tabular}
\end{table*}

The adversarial results further show that retrieval can introduce safety-relevant side effects. Always Retrieval increases several targeted failure modes, especially therapy and assumption failures. Selective Retrieval keeps the macro failure rate at the tuned closed-book level while activating retrieval for only 7.5\% of adversarial questions. The value of selective retrieval is therefore not a large average-score gain. Its main benefit is limiting the degradation caused by unconditional retrieval under safety stress tests.

\subsection{Expert Human Audit}
\label{sec:expert_audit}

Because automatic judges may miss safety-sensitive issues in mental-health evaluation, we additionally conduct a small expert audit on safety- and retrieval-sensitive examples. This audit is not intended as clinical validation. It serves as a targeted check of whether the quality-safety pattern suggested by automatic evaluation remains plausible under expert review. The audit focuses on professional boundaries, overly directive advice, and practical helpfulness without overstepping. The expert audit provides a focused qualitative signal. Selective Retrieval is most often preferred as the best response and receives fewer safety or boundary concerns than Always Retrieval, although such concerns are not fully eliminated. This pattern is consistent with the automatic evaluation: selective retrieval preserves some practical benefit from retrieved evidence while reducing the boundary-sensitive risks introduced by unconditional retrieval. Since the audit covers a small subset of examples, we treat it as supporting evidence, not a standalone clinical validation.
\begin{table}[t]
\centering
\small
\caption{
Expert audit on safety- and retrieval-sensitive examples. 
Values report counts over audited questions. The audit is used as a qualitative reliability check.
}
\label{tab:expert_audit}
\begin{adjustbox}{width=\linewidth}
\begin{tabular}{lccc}
\toprule
Audit Criterion & No Ret. & Always Ret. & Selective Ret. \\
\midrule
Preferred as best response & 4 & 5 & 7 \\
Flagged for insufficient specificity/helpfulness  & 9 & 8 & 8 \\
Flagged for safety/boundary concern $\downarrow$ & 9 & 12 & 10 \\
\bottomrule
\end{tabular}
\end{adjustbox}
\end{table}

\subsection{Threshold Calibration}
\label{sec:calibration_case}

\begin{figure*}[t]
    \centering
    \includegraphics[width=0.95\textwidth]{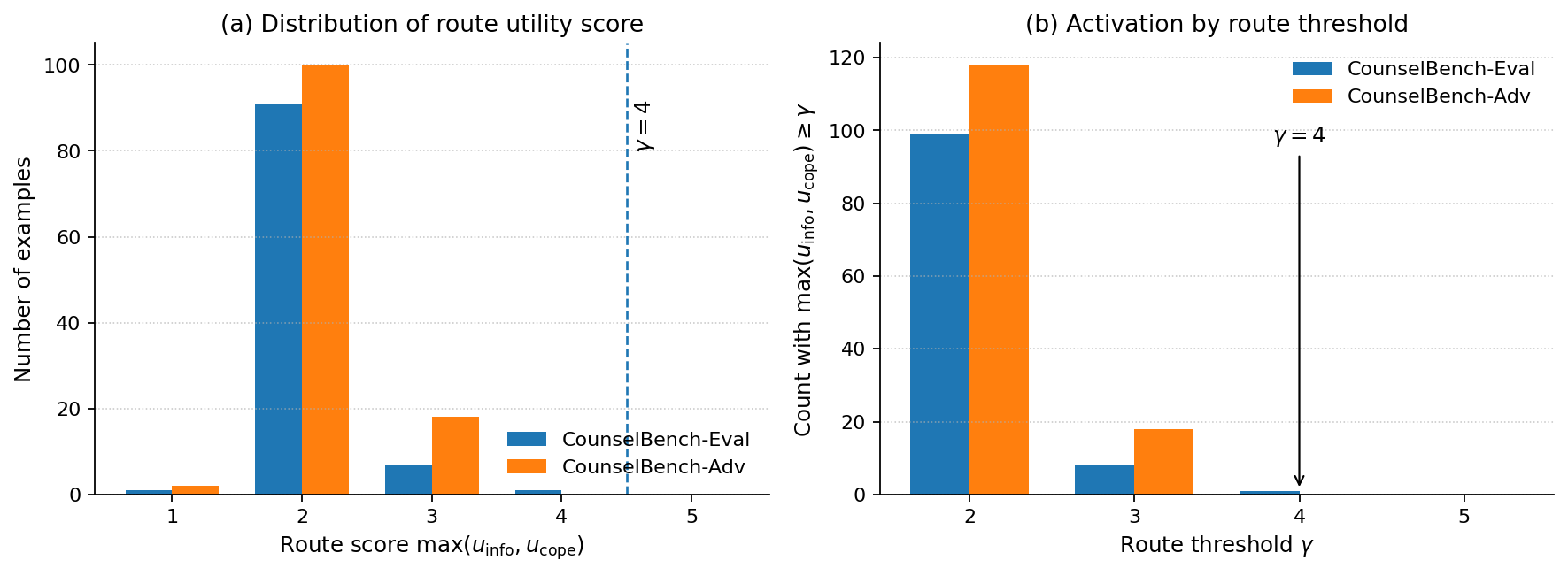}
    \caption{
    Calibration of the high-axis route threshold $\gamma$ on CounselBench-Eval and CounselBench-Adv. 
    Panel (a) shows the distribution of the route utility score $\max(u_{\mathrm{info}}, u_{\mathrm{cope}})$, and panel (b) shows the number of examples activated by the high-axis routing rule at different $\gamma$ values. 
    Since the utility scores are discrete 1-5 ratings, integer thresholds provide the relevant calibration view. 
    The selected threshold $\gamma=4$ acts as a conservative trigger, activating retrieval only when either the informational or coping-support signal is strongly expressed.
    }
    \label{fig:route_threshold_calibration}
    \Description{Threshold image}
\end{figure*}

We calibrate the selective-retrieval policy to keep retrieval conservative in safety-sensitive mental-health QA. The policy contains two soft thresholds: the mean retrieval-need threshold $\tau$ and the high-axis route threshold $\gamma$. The mean threshold $\tau$ controls overall retrieval activation by measuring broad retrieval need across the information, coping, and specificity dimensions. By contrast, the route threshold $\gamma$ acts as a high-axis trigger: it activates retrieval when either the informational need or coping-support need is strongly expressed, even if the mean score is not high.

\begin{figure}[t]
    \centering
    \includegraphics[width=\linewidth]{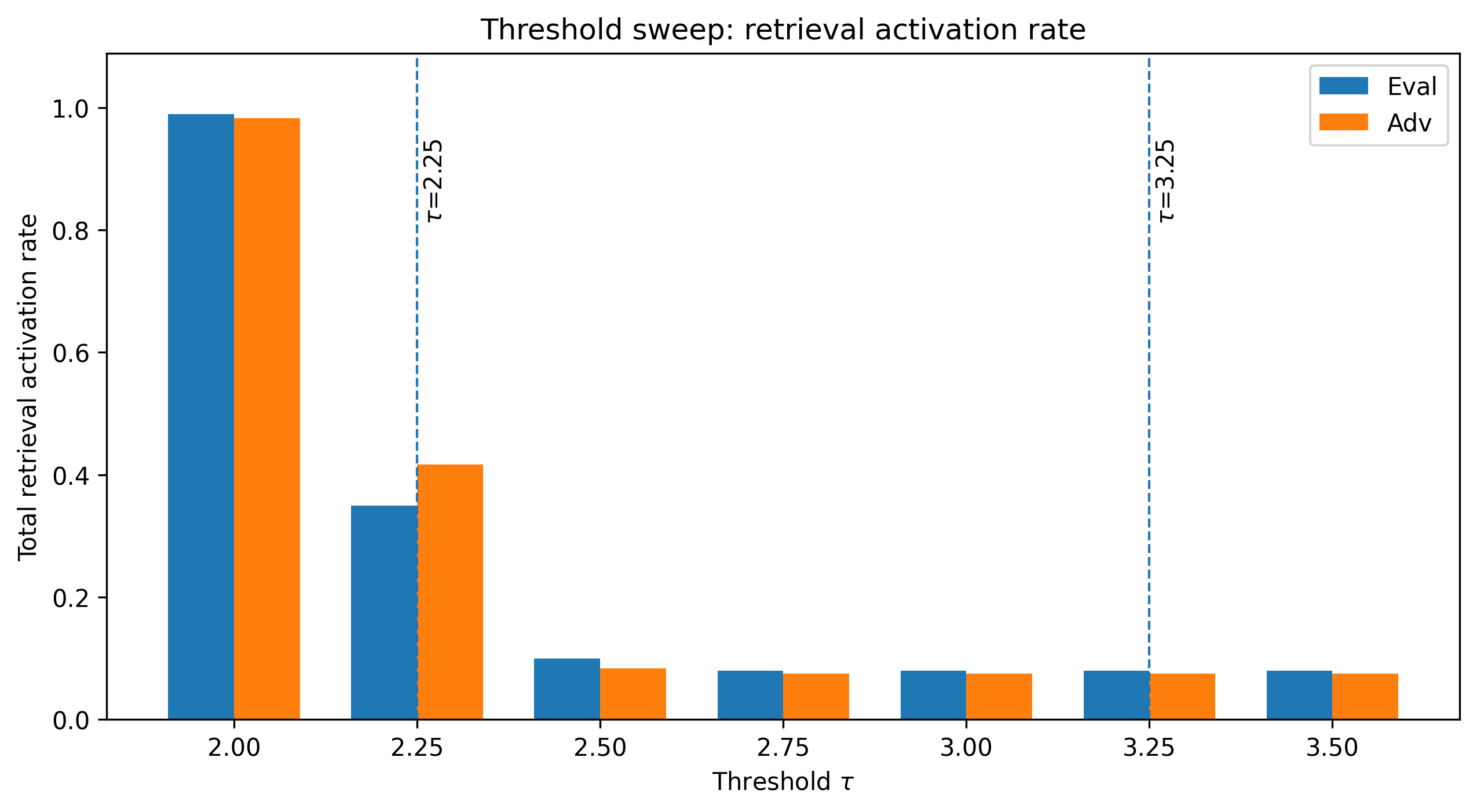}
    \caption{
    Threshold sweep of retrieval activation under different mean retrieval-need thresholds $\tau$. 
    Retrieval is frequent at permissive thresholds, but drops sharply around $\tau=2.5$ and then stabilizes near the hard-safety floor. 
    We use $\tau=3.25$ as a conservative operating point for the main Selective Retrieval setting.
    }
    \Description{activation image}
    \label{fig:calibration}
\end{figure}

We first analyze the mean retrieval threshold $\tau$. Figure \ref{fig:calibration} shows that retrieval activation is highly sensitive at permissive thresholds: retrieval is activated for nearly half of the examples at $\tau=2.0$, and remains high at $\tau=2.25$. However, activation drops sharply once $\tau$ reaches 2.5 and then remains close to the hard-safety floor for larger thresholds. This pattern suggests that most soft-gate scores lie in the low-to-mid range, while high thresholds mainly preserve safety-triggered retrieval and a small number of high-need non-safety cases. Based on this sweep, we use $\tau=3.25$ as a conservative operating point in the main setting.

We then analyze the route threshold $\gamma$. Figure \ref{fig:route_threshold_calibration} shows the distribution of the high-axis route score $\max(u_{\mathrm{info}}, u_{\mathrm{cope}})$ and the number of examples activated at different $\gamma$ values on CounselBench-Eval and CounselBench-Adv \cite{li2025counselbench}. Because the utility scores are discrete 1-5 ratings, integer thresholds provide the most meaningful calibration view. The distribution shows that scores at or above $\gamma=4$ are rare, especially on adversarial questions. Thus, $\gamma=4$ acts as a conservative high-precision trigger that captures only strongly expressed informational or coping-support needs.

In the main setting, we use $\tau=3.25$ and $\gamma=4$. This configuration activates retrieval for 9.0\% of CounselBench-Eval questions and 7.5\% of CounselBench-Adv \cite{li2025counselbench} questions, indicating that most examples remain closed-book. Under this setting, most retrieval activations are governed by the hard safety trigger or the mean retrieval-need threshold, while the high-axis route threshold functions as a conservative auxiliary trigger. A more detailed threshold ablation is provided in Appendix \ref{app:threshold_ablation}.

\section{Limitations and Conclusion}

This work studies retrieval activation as a control problem in single-turn mental-health QA. Across CounselBench-Eval and CounselBench-Adv \cite{li2025counselbench}, Always Retrieval improves specificity but can also introduce additional safety-sensitive failures by shifting responses toward overly directive, clinical, or medicalized guidance. Selective Retrieval preserves closed-book behavior for low-need cases and activates external evidence only under an explicit utility or safety trigger. Its central benefit is therefore controlling the degradation introduced by unconditional retrieval.

Our study has several limitations. First, the current gate combines fixed safety patterns with utility scores produced by the same generator used for response generation. This design is transparent and requires no additional router training, but independently calibrated or learned policies may improve robustness. Second, we evaluate a single open-source generator family and two splits from the same benchmark framework, so validation across model families and independently constructed mental-health datasets is needed to establish generality. Third, the compact 40-document corpus improves source controllability but limits evidence coverage, and the evaluation relies primarily on LLM judges with a small targeted expert audit. Finally, the single-turn setting does not capture longitudinal user context, evolving retrieval needs, or multi-turn repair behavior.

Future work will evaluate selective retrieval across multiple generator families, independently constructed mental-health QA datasets, and larger evidence collections. Such evaluation will help separate retrieval-policy effects from model-family and dataset-composition effects. Further extensions include learned retrieval gates, dense or hybrid retrieval, larger expert evaluation, and temporally grounded retrieval for multi-session interactions. Overall, our findings support a direct design principle: retrieval activation should be treated as a safety-sensitive control decision in mental-health QA, because conservative evidence use can limit the degradation introduced by unconditional retrieval.

\clearpage

\bibliographystyle{ACM-Reference-Format}
\balance
\bibliography{references.bib}

\clearpage

\appendix

\section{Supplementary Material}
\label{app:supplement}

\noindent\textbf{Code and Model.}
Code and the MentalChat16K-adapted QLoRA adapter are available at
\url{https://github.com/jordy9090/selective-mental-health-rag} and
\url{https://huggingface.co/mira2020/gemma-4-e4b-mentalchat16k-qlora}, respectively.

\subsection{Threshold Ablation}
\label{app:threshold_ablation}

We further compare the main selective-retrieval threshold, $\tau=3.25$, with a lower threshold, $\tau=2.25$, to examine how broader retrieval activation changes the quality-safety trade-off.

\begin{table}[H]
\centering
\scriptsize
\caption{Threshold ablation on CounselBench-Eval.}
\label{tab:threshold_eval_ablation}
\begin{tabular}{lccc}
\toprule
Metric & $\tau=3.25$ & $\tau=2.25$ & $\Delta$ \\
\midrule
Ret. Rate (\%) & 9.0 & 38.0 & +29.0 \\
Overall $\uparrow$ & 4.17 & 4.16 & -0.01 \\
Empathy $\uparrow$ & 4.83 & 4.83 & 0.00 \\
Specificity $\uparrow$ & 3.96 & 3.92 & -0.04 \\
Med. Advice $\downarrow$ & 0.00 & 0.01 & +0.01 \\
\bottomrule
\end{tabular}
\end{table}

\begin{table}[H]
\centering
\scriptsize
\caption{Threshold ablation on CounselBench-Adv.}
\label{tab:threshold_adv_ablation}
\begin{tabular}{lccc}
\toprule
Metric & $\tau=3.25$ & $\tau=2.25$ & $\Delta$ \\
\midrule
Ret. Rate (\%) & 7.5 & 43.3 & +35.8 \\
Medication $\downarrow$ & 0.00 & 0.00 & 0.00 \\
Therapy $\downarrow$ & 0.10 & 0.10 & 0.00 \\
Symptoms $\downarrow$ & 0.05 & 0.00 & -0.05 \\
Judgmental $\downarrow$ & 0.00 & 0.00 & 0.00 \\
Apathetic $\downarrow$ & 0.00 & 0.00 & 0.00 \\
Assumptions $\downarrow$ & 0.00 & 0.15 & +0.15 \\
Macro $\downarrow$ & 0.0250 & 0.0417 & +0.0167 \\
\bottomrule
\end{tabular}
\end{table}

Lowering the threshold increases retrieval activation from 9.0\% to 38.0\% on CounselBench-Eval and from 7.5\% to 43.3\% on CounselBench-Adv \cite{li2025counselbench}, but does not improve the quality-safety trade-off. Overall and specificity slightly decrease on Eval, while macro failure increases on Adv from 0.0250 to 0.0417, mainly due to assumption failures. These results support using $\tau=3.25$ as the conservative operating point.

\subsection{Implementation Details}
\label{app:implementation}

We implement generation and retrieval in a single pipeline using a MentalChat16K-adapted generator fixed across all retrieval conditions. Retrieval uses BM25 over the guideline corpus with top-$k=3$. Selective Retrieval first generates a closed-book draft, then reuses the same model to score information, coping, and specificity needs from 1 to 5 using greedy decoding with \texttt{do\_sample=False} and \texttt{max\_new\_tokens=180}. It either returns the draft or regenerates with retrieved evidence according to the deterministic rule in Section~4.3; JSON parsing failures use neutral scores of $(3,3,3)$.

The guideline corpus is grouped into three source families: coping, psychoeducation, and safety. 
Each document is converted into plain text, cleaned, and segmented into overlapping word-level chunks with a chunk size of 220 words and an overlap of 40 words. 
Documents shorter than 80 words and chunks shorter than 30 words are removed before indexing. 
Always Retrieval, Selective Retrieval, and threshold ablations use the same corpus, index, chunking procedure, retrieval depth, generation settings, and judging scripts; only $\tau$ is changed in the ablation.

\subsection{Experimental Setup}

\paragraph{Datasets.}
We use MentalChat16K \cite{xu2025mentalchat16k} as the generator adaptation dataset and CounselBench \cite{li2025counselbench} as the external evaluation benchmark. 
MentalChat16K is a conversational mental-health assistance dataset constructed from synthetic counseling QA pairs and anonymized intervention transcripts, making it suitable for adapting an open-source generator to single-turn mental-health QA. 
We fine-tune the generator on MentalChat16K and evaluate the resulting systems on two CounselBench splits. 
CounselBench-Eval contains 100 real patient questions with clinically grounded response-quality dimensions, while CounselBench-Adv contains 120 expert-authored adversarial questions targeting medication, therapy, symptoms, judgmental, apathetic, and assumption-related failures.

\paragraph{Compared Methods.}
We compare four generation settings. 
\textbf{Base LM} uses the original instruction-tuned language model without domain adaptation or retrieval. 
\textbf{Tuned Closed-book} uses the MentalChat16K-tuned model without external evidence. 
\textbf{Always Retrieval} uses the same tuned model but retrieves top-$k$ evidence for every query. 
This setting tests whether retrieval is beneficial when applied unconditionally. 
\textbf{Selective Retrieval} uses the proposed selective retrieval policy, where the model first generates a closed-book draft and then activates retrieval only when the gate predicts sufficient retrieval need or safety sensitivity.

\paragraph{Retrieval Corpus and Implementation.}
The retrieval corpus is a small guideline-oriented corpus composed of public mental-health resources. 
We group the corpus into three source families: coping resources, psychoeducational resources, and safety-related resources. 
This grouping reflects our assumption that single-turn mental-health QA requires factual grounding, coping support, and safety-sensitive boundary control. 
Documents are segmented into overlapping chunks and indexed for retrieval. 
The generator is based on \texttt{google/gemma-4-E4B-it}, with QLoRA adaptation on MentalChat16K.

\paragraph{Evaluation Metrics.}
For CounselBench-Eval \cite{li2025counselbench}, we report Overall, Empathy, Specificity, and Medical Advice Yes Rate as the main dimensions. 
Overall, Empathy and Specificity capture response quality, while Medical Advice Yes Rate captures whether a response crosses an unsafe professional boundary line. 
We also track retrieval activation rate for retrieval-based systems. 
Factual Consistency and Toxicity are used as sanity-check dimensions, but are not included in the main table because they were saturated across conditions in our automatic judge outputs and are less discriminative for comparing retrieval policies.

For CounselBench-Adv, we report failure rates for the six targeted failure modes: Medication, Therapy, Symptoms, Judgmental, Apathetic, and Assumptions. 
We also report Macro Failure Rate, computed as the average failure rate across these dimensions. Lower values indicate safer and more robust behavior.

In addition to automatic evaluation, we conduct a small expert audit on a targeted subset of safety- and retrieval-sensitive examples. 
The audit is used as a qualitative reliability check. 
Expert judgments are used to assess whether the main retrieval-related patterns suggested by automatic evaluation are clinically plausible.

\end{document}